\documentclass[journal]{IEEEtran}
\usepackage{amsmath,amsfonts}
\usepackage{bbm}
\usepackage{xcolor}
\usepackage{algorithmic}
\usepackage{algorithm}
\usepackage{booktabs}
\usepackage{array}
\usepackage[caption=false,font=normalsize,labelfont=sf,textfont=sf]{subfig}
\usepackage{textcomp}
\usepackage{stfloats}
\usepackage{url}
\usepackage{verbatim}
\usepackage{graphicx}
\usepackage[T1]{fontenc}
\graphicspath{ {./figures/} }

\begin{document}

\title{Imagination-Augmented Hierarchical \\
Reinforcement Learning for Safe and Interactive \\
Autonomous Driving in Urban Environments}

\author{Sang-Hyun Lee, \IEEEmembership{Member,~IEEE}, Yoonjae Jung, \IEEEmembership{Member,~IEEE}, and Seung-Woo Seo, \IEEEmembership{Member,~IEEE}
\thanks{Manuscript received ----- 00, 2023; revised ------ 00, 2024.
This research was supported by the Challengeable Future Defense Technology Research and Development Program through the Agency For Defense Development(ADD) funded by the Defense Acquisition Program Administration(DAPA) in 2023(No.915027201), the Institute of New Media and Communications, the Institute of Engineering Research, and the Automation and Systems at Seoul National University. (Corresponding author: Seung-Woo Seo)} 
\thanks{Sang-Hyun Lee, Yoonjae Jung, and Seung-Woo Seo are with the Department of Electrical and Computer Engineering, Seoul National University, Korea (e-mail: slee01@snu.ac.kr; yujs30@snu.ac.kr; sseo@snu.ac.kr)}
\thanks{This work has been submitted to the IEEE for possible publication. Copyright may be transferred without notice, after which this version may no longer be accessible.}
}

\maketitle

\begin{abstract}
Hierarchical reinforcement learning (HRL) incorporates temporal abstraction into reinforcement learning (RL) by explicitly taking advantage of hierarchical structure. Modern HRL typically designs a hierarchical agent composed of a high-level policy and low-level policies. The high-level policy selects which low-level policy to activate at a lower frequency and the activated low-level policy selects an action at each time step. Recent HRL algorithms have achieved performance gains over standard RL algorithms in synthetic navigation tasks. However, we cannot apply these HRL algorithms to real-world navigation tasks. One of the main challenges is that real-world navigation tasks require an agent to perform safe and interactive behaviors in dynamic environments. In this paper, we propose imagination-augmented HRL (IAHRL) that efficiently integrates imagination into HRL to enable an agent to learn safe and interactive behaviors in real-world navigation tasks. Imagination is to predict the consequences of actions without interactions with actual environments. The key idea behind IAHRL is that the low-level policies imagine safe and structured behaviors, and then the high-level policy infers interactions with surrounding objects by interpreting the imagined behaviors. We also introduce a new attention mechanism that allows our high-level policy to be permutation-invariant to the order of surrounding objects and to prioritize our agent over them. To evaluate IAHRL, we introduce five complex urban driving tasks, which are among the most challenging real-world navigation tasks. The experimental results indicate that IAHRL enables an agent to perform safe and interactive behaviors, achieving higher success rates and lower average episode steps than baselines.
\end{abstract}

\section{Introduction}
\IEEEPARstart{H}{ierarchical} reinforcement learning (HRL) \cite{parr1997reinforcement, dietterich2000hierarchical, stolle2002learning}, which incorporates the idea of different temporal abstraction levels into RL, has been actively researched in various domains. Modern HRL typically designs a hierarchical agent that consists of two parts: a high-level policy and low-level policies. The high-level policy determines which low-level policy to activate over a longer time scale, and the activated low-level policy outputs an action at every time step. Recent works on HRL have achieved noticeable performance improvement in synthetic navigation tasks, including mazes \cite{eysenbach2019search, gieselmann2021planning, wohlke2021hierarchies} and racing games \cite{vezhnevets2017feudal, hao2023skill}. However, applying them to real-world navigation tasks poses significant challenges. First, real-world navigation tasks require an agent to interact with surrounding objects in rapidly changing environments. An agent has no control over surrounding objects, and both can affect each other. Second, an agent in real-world navigation tasks should perform safety-aware behaviors following task-specific rules. Specifying a reward function manually for these behaviors is not straightforward. Third, real-world navigation tasks are long-horizon and have complex structures with diverse semantic objects. This requires an agent to perform consistent and structured behaviors to explore environments. Unfortunately, developing an HRL agent that can handle these challenges remains an open problem \cite{kiran2021deep, zhu2021survey}.

\begin{figure}[t]
\begin{center}
\centerline{\includegraphics[width=1.0\columnwidth]{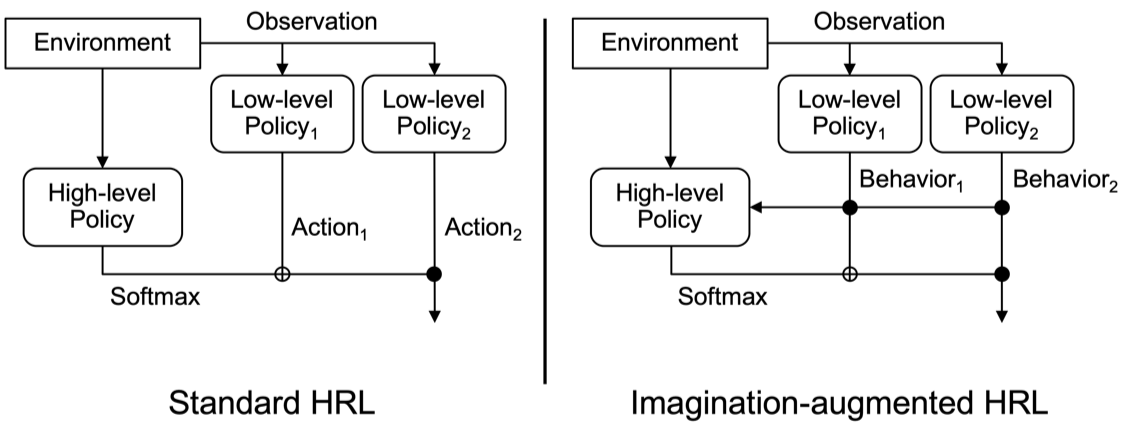}}
\vskip -0.05in
\caption{Overview of IAHRL. Both standard HRL and IAHRL temporally decompose a given task with their hierarchical structures. However, unlike standard HRL, IAHRL allows a high-level policy to interpret the output of low-level policies. Furthermore, the low-level policies in IAHRL generate distinct behaviors rather than single-step actions.
}
\label{fig1:imagination-augmented}
\end{center}
\vskip -0.15in
\end{figure}

In this paper, we propose imagination-augmented HRL (IAHRL) that can address the above challenges in real-world navigation tasks. IAHRL efficiently integrates the concept of imagination into HRL to enable an agent to learn safe and interactive behaviors. Imagination, which is known to play an important role in long-term reasoning \cite{silver2016mastering, anthony2017thinking, racaniere2017imagination, hafner2019dream}, is to reason about the future by predicting outcomes of an action without interaction with actual environments. The key idea behind IAHRL is that the low-level policies imagine safe and structured behaviors, and then the high-level policy infers interaction with surrounding objects by interpreting the imagined behaviors. Figure \ref{fig1:imagination-augmented} describes an overview of IAHRL. Both standard HRL and IAHRL utilize a hierarchical structure to design an agent. However, IAHRL has two distinctions as follows: 1) IAHRL allows the high-level policy to interpret the outputs of the low-level policies to determine which low-level policy is activated, and 2) the low-level policies are designed to output behaviors to be followed, not just sing-step actions.

We also introduce a new attention mechanism that can enable our high-level policy to be permutation-invariant to the order of surrounding objects and to prioritize our agent over them. This can help our agent to perform more stable behaviors than previous works that naively train an agent with the stack of surrounding object features, which are susceptible to their order \cite{qiao2020hierarchical, qiao2021behavior, li2021safe}. Furthermore, our agent is also robust to imagination errors, as IAHRL trains it to learn how to interpret imagined behaviors rather than relying on them blindly. Our low-level policies are implemented with an optimization-based behavior planner to imagine safety-aware behaviors following task-specific rules. Each low-level policy is designed to imagine behaviors for a different high-level action, such as lane following and lane changing. This design choice enables our hierarchical agent to explore complex environments with coherent and structured behaviors efficiently.

To evaluate our algorithm, we introduce five autonomous driving tasks in urban environments. Urban autonomous driving, which has been spotlighted with its massive potential for decades \cite{thorpe19971997, tan1998demonstration, dickmanns2007dynamic}, is one of the most challenging real-world navigation tasks. There are diverse traffic objects and rules in urban driving environments, and it is impractical to handle them with conventional autonomous driving systems based on hand-crafted algorithms \cite{urmson2008autonomous, leonard2008perception, ziegler2014making, broggi2015proud}. The main contributions of our work are three-fold: 1) we introduce IAHRL that efficiently integrates imagination into HRL to enable an agent to perform safe and interactive behaviors in real-world navigation tasks, 2) we propose a new permutation-invariant attention mechanism that allows our high-level policy to infer interactions with surrounding objects from behaviors imagined by low-level policies, and 3) we introduce five complex urban autonomous driving tasks using the open-source urban driving simulator CARLA \cite{dosovitskiy2017carla}. Experimental results indicate that our hierarchical agent performs safety-aware behaviors and interacts successfully with surrounding vehicles, achieving higher success rates and fewer average episode steps than baselines.

\section{Related Works}
IAHRL can be interpreted as a navigation algorithm that takes observations as input and infers behaviors to be followed. Conventional navigation algorithms, such as time-to-collision (TTC) \cite{lee1976theory} and dynamic window approach (DWA) \cite{fox1997dynamic}, are implemented based on hand-crafted functions. The DARPA Urban Challenge \cite{buehler2009darpa}, a milestone event held in 2007, showed that such conventional algorithms could allow autonomous vehicles to follow traffic rules and complete a 60-mile route in restricted driving environments \cite{leonard2008perception, montemerlo2008junior}. The winner of this event \cite{urmson2008autonomous}, called BOSS, successfully used a slot-based approach that evaluates a set of available slots to determine when to merge into target lanes or enter intersections \cite{baker2008traffic}. However, applying these conventional algorithms to real-world navigation tasks where diverse and complex scenarios should be handled is impractical, as hand-crafted functions depend heavily on domain knowledge and consist of rules restricted to specific scenarios. IAHRL can overcome this limitation of conventional algorithms by learning navigation behaviors without hand-crafted rules.

Learning-based navigation algorithms broadly fall into two categories: imitation learning (IL) and RL algorithms. IL algorithms seek to learn navigation policies directly from expert demonstrations. ALVINN, one of the earliest IL navigation algorithms, demonstrated that a shallow neural network can be trained to perform lane-following behaviors in rural driving environments \cite{pomerleau1988alvinn}. Inspired by this pioneering work, Bojarski et al. \cite{bojarski2016end} used a deep convolutional neural network to obtain a robust navigation system for driving on public roads. Bansal et al. \cite{bansal2018chauffeurnet} introduced ChauffeurNet that trains an agent from expert demonstrations augmented with synthesized perturbations. This allows the agent to experience nonexpert behaviors such as collisions. Pini et al. \cite{pini2023safe} presented SafePathNet that uses a Mixture of Experts (MoE) approach to predict multi-modal trajectories of surrounding and ego vehicles. They showed that the predicted trajectory with the minimum cost allows the ego vehicle to handle several urban driving scenarios. While these impressive works have led to further research in various directions, IL algorithms are known to generalize poorly in complex tasks due to compounding errors caused by covariate shifts \cite{ross2010efficient, ross2011reduction}. Furthermore, collecting useful expert demonstrations is prohibitively expensive in most real-world settings. IAHRL circumvents these limitations of IL algorithms by training an agent to maximize reward signals obtained from interactions with environments.

In contrast to IL algorithms, RL algorithms can learn navigation strategies without human-labeled data or expert demonstrations. Lillicrap et al. \cite{lillicrap2015continuous} introduced DDPG that combines the deterministic policy gradient with several innovations in DQN \cite{mnih2015human}. Their experimental results empirically demonstrated that DDPG enabled an agent to complete a circuit around a track in the racing car simulator TORCS \cite{wymann2000torcs}. Kendall et al. \cite{kendall2019learning} demonstrated that a full-sized vehicle can be trained with RL to handle lane-following scenarios. While these algorithms do not explicitly take advantage of hierarchical structures, several HRL algorithms have demonstrated that leveraging hierarchical structures leads to noticeable performance gains over flat RL algorithms on long-horizon navigation tasks. Vezhnevets et al. \cite{vezhnevets2017feudal} introduced FeUdal Networks (FuNs) that utilize learned abstract goals to communicate between high- and low-level modules. Their experimental results empirically demonstrated that FuNs can handle a challenging racing game requiring long-term reasoning to play well. Eysenbach et al. \cite{eysenbach2019search} proposed SoRB that uses planning via graph search as a fixed high-level policy in a hierarchical agent. This allows an agent to decompose a navigation task of reaching a distant goal into a sequence of easier goal-reaching tasks. While the experimental results of these HRL algorithms support our design choice to leverage hierarchical structure to address long-term navigation tasks, these algorithms only focused on learning navigation behaviors in static environments without considering interaction with surrounding objects.

Our work presented here belongs to the class of HRL and focuses on learning safe and interactive behaviors in dynamic environments, which is required to solve real-world navigation tasks. Similar to our work, several recent works on HRL proposed impressive algorithms to learn interactive behaviors in dynamic environments \cite{shalev2016safe, gao2022cola, wang2023efficient, gu2023safe}. Gao et al. \cite{gao2022cola} introduced Cola-HRL that trains the high-level policy to generate a goal under the Frenet coordinate system and utilizes the MPC-based low-level policy to output control commands to follow the generated goal. Wang et al. \cite{wang2023efficient} presented ASAP-RL that uses the set of parameterized motion skills learned from expert demonstrations as the action space of the high-level policy. While both works showed that their HRL algorithms can handle diverse urban driving scenarios, they cannot guarantee the safety of their driving behaviors. Gu et al. \cite{gu2023safe} proposed Safari that leverages state-based constraints to ensure the safety of goals generated by the high-level policy. However, Safari is verified only in scenarios in highway environments. Our work demonstrates that IAHRL can enable an agent to perform safe and interactive behaviors in diverse urban driving scenarios.

IAHRL is most closely related to H-CtRL introduced by Li et al. \cite{li2021safe}. H-CtRL ensures the safety of an agent by implementing low-level policies with an optimization-based behavior planner. While both IAHRL and H-CtRL demonstrated impressive results on diverse driving scenarios, IAHRL has two clear distinctions as follows: 1) IAHRL integrates imagination into HRL, allowing the high-level policy to infer interactions with surrounding objects by interpreting future behaviors imagined by the low-level policies, and 2) IAHRL is permutation-invariant to the order of surrounding objects and prioritizes our agent over them. In contrast, H-CtRL infers interactions based on the current state without imagination, and the current state is defined as the stack of features of surrounding objects, which is susceptible to their order. Our experimental results show that these distinctions of IAHRL contribute to better performance gain over H-CtRL.

Incorporating imagination into RL has demonstrated promising results on various tasks \cite{silver2016mastering, racaniere2017imagination, hafner2019dream, li2023imagination}. Silver et al. \cite{silver2016mastering} introduced AlphaGo, the Go-playing agent, that leverages an environmental model to infer the possible future for each action and determines the action that leads to the best future. AlphaGo defeated the best professional Go player, which was a milestone demonstration verifying the massive potential of imagination. However, this work does not address imagination errors caused by imperfect model approximations. To handle this issue, Weber et al. \cite{racaniere2017imagination} presented an imagination-augmented RL that is robust against model imperfection by learning to interpret imagined behaviors. Hafner et al. \cite{hafner2019dream} proposed Dreamer that can learn farsighted behaviors using latent imagination based on a pre-trained world model. Li et al. \cite{li2023imagination} demonstrated that interpreting imagined behaviors enables an RL agent to address variable speed limit optimization problems even with imperfect traffic flow models. Inspired by these previous works that integrate imagination into RL, we incorporate imagination into HRL, enabling an agent to perform safe and interactive behaviors in real-world navigation tasks. Our work allows the agent to imagine safe and structured behaviors by implementing low-level policies with an optimization-based behavior planner.

The concept of the attention mechanism has been utilized in various research fields \cite{devlin2018bert, radford2019language, chen2021decision}. The Transformer, introduced by Vaswani et al. \cite{vaswani2017attention}, is the first sequence transduction model relying entirely on attention. This achieved a substantial performance gain over previous works based on recurrent or convolutional structures. Dosovitski et al. \cite{dosovitskiy2020image} interpreted the image as a sequence of patches so that the attention mechanism can capture the meaningful context representation of images. Prakash et al. \cite{prakash2021multi} proposed TransFuser that uses the attention mechanism to integrate representations from different modalities such as image and LiDAR. Tang et al. \cite{tang2021sensory} introduced permutation-invariant RL agents that leverage the new self-attention mechanism to handle an unordered observation set. Their experimental results show that the permutation-invariant agents are more robust and generalize better to unseen tasks. In this work, we introduce a new permutation-invariant attention mechanism that can interpret the imagined behaviors of our agent and surrounding objects and prioritize our agent over surrounding objects.

\section{Preliminaries}
\subsection{Markov Decision Process (MDP)}
We formally describe an environment using a Markov decision process (MDP), represented as the tuple $(S, A, P, R, \rho_0, \gamma, T)$. $S$ refers to the set of states, $s$, $A$ refers to the set of actions $a$, and $P: S \times A \times S \rightarrow [0,1]$ is the state transition model. $R: S \times A \rightarrow \mathbb{R}$ denotes the reward function that outputs scalar feedback, $r$, called reward. $\rho_0: S \rightarrow [0,1]$ is the initial state distribution, $\gamma$ is the discount factor, and $T$ is the horizon. An agent takes action sampled from a policy, $\pi: S \times A \rightarrow [0,1]$, which maps states to a probability distribution over actions. 

MDP can be generalized as a partially observable MDP (POMDP) where we assume that an agent cannot directly access the underlying states of the environment. POMDP is a useful framework for taking into account uncertainties in decision-making problems.  However, since our work focuses on efficiently integrating imagination into HRL to ensure safe and interactive behaviors in navigation tasks, we assume an agent can directly observe the underlying states, and leave extending our work to POMDP for future work.

RL is an approach for finding the optimal policy $\pi^*$ that maximizes the expected cumulative rewards when the dynamics of the environment is unknown. One of the main elements in RL is the state-action value function $Q^\pi(s,a)$, which can be formalized as follows: 
\begin{equation*} \label{q_function}
Q^{\pi}(s, a)=\mathbb{E}_{\pi}\Big[\textstyle{\sum^\infty_{i=t}}\gamma^{i-t} r_{i+1}|S_t=s, A_t=a\Big].
\end{equation*}
The state-action value function represents the expected discounted sum of rewards obtained when an agent takes the action $a$ in the state $s$ and follows the policy $\pi$.

\subsection{Self-attention}
The self-attention mechanism effectively captures the interdependencies within a variable number of elements of an input sequence. It can be formalized as $m=\sigma(\mathcal{Q}\mathcal{K}^T)\mathcal{V}$, where $\sigma$ is a softmax function and $\mathcal{Q}, \mathcal{K},$ and $\mathcal{V}$ are Query, Key, and Value matrices. These matrices are commonly defined as functions of an input sequence. The dot product between Query and Key, referred to as the attention score, estimates the relations between them, and the softmax function transforms the score into a normalized weight vector. Thus, the output of the attention mechanism is the weighted average of value features. The attention output has a permutation equivariance (PE) property. This means that if the order of the input sequence changes, the order of the attention output also changes. In contrast, our attention mechanism introduced in this work is permutation invariant to the order of the input sequence and can prioritize some elements more than others. We use our attention mechanism to implement the high-level policy.

\begin{figure*}[t]
\begin{center}
\centerline{\includegraphics[width=0.70\textwidth]{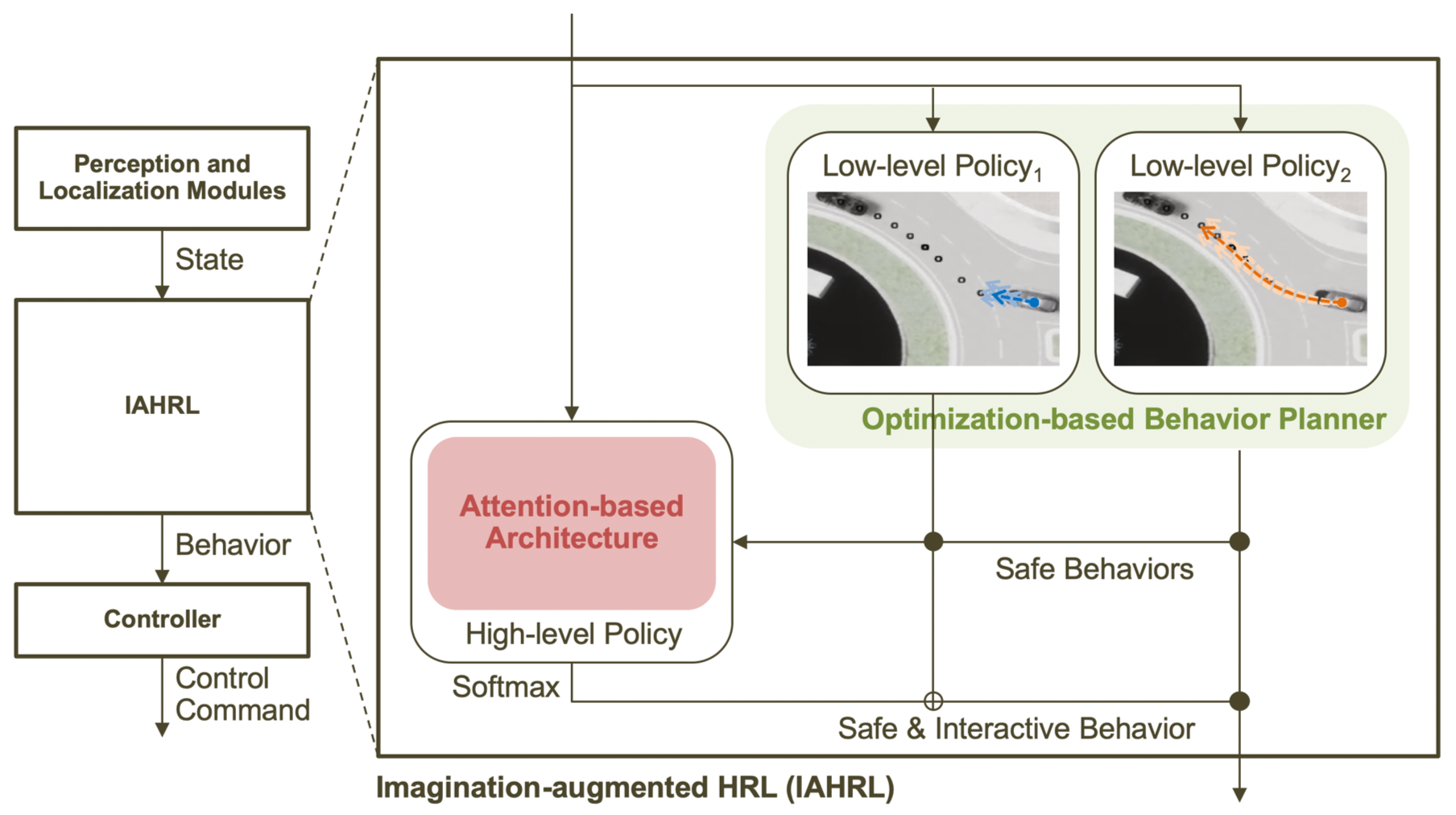}}
\caption{Structure of IAHRL. IAHRL takes the localization and perception results as input and outputs a behavior to be transmitted to the control module. The low-level policies implemented with an optimization-based behavior planner imagine safe and structured behaviors for each high-level action. The high-level policy implemented with a new attention mechanism infers interactions with surrounding objects from the behaviors imagined by the low-level policies, and selects the low-level policy that imagines the most interactive behavior.}
\label{fig2:overview}
\end{center}
\vskip -0.2in
\end{figure*}

\section{Imagination-Augmented \\ Hierarchical Reinforcement Learning}
Our work aims to enable an agent to perform safe and interactive behaviors in real-world navigation tasks. To achieve this goal, IAHRL efficiently combines HRL with imagination that is to reason about the future by predicting the consequences of an action without interaction with actual environments \cite{silver2016mastering, racaniere2017imagination}. The key idea behind IAHRL is that the low-level policies imagine safe and structured behaviors, and then the high-level policy interprets interaction with surrounding objects based on the imagined behaviors. This idea is based on the hypothesis that imagined behaviors are critical to reason about interactions in the future. IAHRL contrasts typical prior works that train an agent to infer interactions from previous or current states. In addition, we would like to note that IAHRL is robust to imagination errors, as it learns how to interpret the imagination results rather than relying on them blindly.

Following widely adopted modular pipelines in the industry, we assume that IAHRL has access to localization, perception, and control modules. As shown in Figure \ref{fig2:overview}, IAHRL takes the results from the localization and perception modules as input and outputs a behavior to be transmitted to the control module. The localization results involve the position of our agent, and the perception results involve predicted behaviors of surrounding objects. The control module generates atom actions, such as throttle, brake, and steering wheel angle (SWA), to follow the behavior imagined by IAHRL.

\begin{table}[t]
\caption{Notations for Main Functions and Variables \label{tab:functions_variables}}
\begin{center}
\resizebox{0.95\columnwidth}{!}{
\begin{tabular}{c c}
\toprule
\qquad SYMBOL \qquad & \qquad DESCRIPTION \qquad \\
\midrule
\qquad $T$ \qquad & \qquad Episode Horizon \qquad \\
\vspace{0.2mm}
\qquad  $H$ \qquad & \qquad High-Level Action Horizon \qquad \\
\vspace{0.2mm}
\qquad  $K$ \qquad & \qquad Imagination Horizon \qquad \\
\vspace{0.2mm}
\qquad $s_t$ \qquad & \qquad State \qquad \\
\vspace{0.2mm}
\qquad $z_t$ \qquad & \qquad High-Level Action \qquad \\
\vspace{0.2mm}
\qquad $a_t$ \qquad & \qquad Low-Level Action \qquad \\
\vspace{0.2mm}
\qquad $\pi_\phi(z_t|s_t)$ \qquad & \qquad High-Level Policy \qquad \\
\vspace{0.2mm}
\qquad $Q^\pi_\theta(s_t,z_t)$ \qquad & \qquad High-Level State-Action Value Function \qquad \\
\vspace{0.2mm}
\qquad $Q^\pi_{\bar{\theta}}(s_t,z_t)$ \qquad & \qquad Target Function of $Q^\pi_\theta(s_t,z_t)$ \qquad \\
\vspace{0.2mm}
\qquad $V^\pi(s_t)$ \qquad & \qquad High-Level State Value Function \qquad \\
\vspace{0.2mm}
\qquad $\pi_{z_i}(a_t|s_t)$ \qquad & \qquad Low-Level Policy \qquad \\
\midrule
\qquad $s(t)$ \qquad & \qquad Covered Arc Length (Frenet Frame) \qquad \\
\vspace{0.2mm}
\qquad $d(t)$ \qquad & \qquad Perpendicular Offset (Frenet Frame) \qquad \\
\vspace{0.2mm}
\qquad $B_{\text{lon}}$ \qquad & \qquad Longitudinal Candidate Behaviors \qquad \\
\vspace{0.2mm}
\qquad $B_{\text{lat}}$ \qquad & \qquad Lateral Candidate Behaviors \qquad \\
\vspace{0.2mm}
\qquad $C_{\text{lon}}$ \qquad & \qquad Longitudinal Cost \qquad \\
\vspace{0.2mm}
\qquad $C_{\text{lat}}$ \qquad & \qquad Lateral Cost \qquad \\
\vspace{0.2mm}
\qquad $C_{\text{tot}}$ \qquad & \qquad Weighted Total Cost \qquad \\
\midrule
\qquad $s_t^{i}$ \qquad & \qquad Attention Input \qquad \\
\vspace{0.2mm}
\qquad $\eta$ \qquad & \qquad Random Seed Vector \qquad \\
\vspace{0.2mm}
\qquad $m_t^{i}$ \qquad & \qquad Attention Output \qquad \\
\vspace{0.2mm}
\qquad $\zeta_t^{\text{ego}_i}$ \qquad & \qquad Imagined Behavior of an Agent \qquad \\
\vspace{0.2mm}
\qquad $\zeta_t^{i}$ \qquad & \qquad Imagined Behavior of Surrounding Object \qquad \\
\vspace{0.2mm}
\qquad $\mathcal{Q}(\zeta_t^{\text{ego}_i}, \eta)$ \qquad & \qquad Query Matrix \qquad \\
\vspace{0.2mm}
\qquad $\mathcal{K}(s_t^i)$ \qquad & \qquad Key Matrix \qquad \\
\vspace{0.2mm}
\qquad $\mathcal{V}(s_t^i)$ \qquad & \qquad Value Matrix \qquad \\
\vspace{0.2mm}
\qquad $W_q$ \qquad & \qquad Projection Network for $\mathcal{Q}(\zeta_t^{\text{ego}_i}, \eta)$ \qquad \\
\vspace{0.2mm}
\qquad $W_k$ \qquad & \qquad Projection Network for $\mathcal{K}(s_t^i)$ \qquad \\
\vspace{0.2mm}
\qquad $W_v$ \qquad & \qquad Projection Network for $\mathcal{V}(s_t^i)$ \qquad \\
\midrule
\qquad $\bar{r}(s_t,z_t)$ \qquad & \qquad High-Level Reward Function \qquad \\
\vspace{0.2mm}
\qquad $\alpha$ \qquad & \qquad Temperature \qquad \\
\vspace{0.2mm}
\qquad $\bar{H}$ \qquad & \qquad Target Entropy \qquad \\
\vspace{0.2mm}
\qquad $\delta$ \qquad & \qquad Learning Rate \qquad \\
\vspace{0.2mm}
\qquad $\omega$ \qquad & \qquad Target Smoothing Coefficient \qquad \\
\vspace{0.2mm}
\qquad $J_{\pi}(\phi)$ \qquad & \qquad Objective Function of $\pi_\phi(z_t|s_t)$ \qquad \\
\vspace{0.2mm}
\qquad $J_{Q^\pi}(\theta)$ \qquad & \qquad Objective Function of $Q^\pi_\theta(s_t,z_t)$ \qquad \\
\vspace{0.2mm}
\qquad $J(\alpha)$ \qquad & \qquad Objective Function of $\alpha$ \qquad \\
\vspace{0.2mm}
\qquad $D$ \qquad & \qquad Replay Buffer \qquad \\
\bottomrule
\end{tabular}
}
\end{center}
\vskip -0.15in
\end{table}

Figure \ref{fig2:overview} also illustrates the structure of IAHRL, including the relationship between the high-level policy $\pi(z|s)$ and the low-level policies $\pi_{z_i}(a|s)$. The low-level policies take the localization and perception results as input and imagine safe and structured behaviors for each high-level action, such as lane changing, following, and stopping. 
The high-level policy takes as inputs the localization and perception results as well as the behaviors imagined by the low-level policies and activates the low-level policy that imagines the most interactive behavior. Note that the high-level policy determines which low-level policy to activate over a longer time scale, H, and the activated low-level policy imagines a behavior at every time step within H steps. In the remainder of this section, we describe in detail how our high-level and low-level policies are designed to perform safe and interactive behaviors. The notations for the main functions and variables used to describe IAHRL are listed in Table \ref{tab:functions_variables}.

\subsection{Safe and Structured Behavior Imagination via \\ Low-Level Policies }
\begin{figure}[t]
\begin{center}
\centerline{\includegraphics[width=0.8\columnwidth]{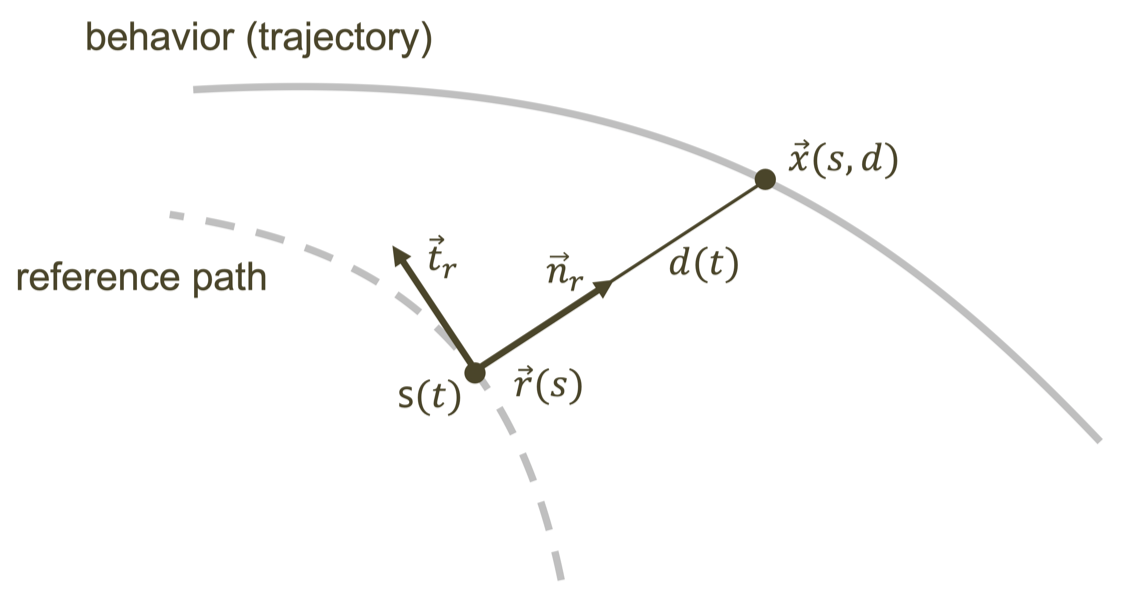}}
\caption{Behavior representation in a Frenet frame. The behaviors imagined by our low-level policies are represented in the Frenet frame, which is defined by the tangential vector $\vec{t}_r$ and the normal vector $\vec{n}_r$.
}
\label{fig3:frenet_frame}
\end{center}
\vskip -0.15in
\end{figure}

To imagine a safe and structured behavior for each high-level action, we implement the low-level policies with an optimization-based behavior planner. We use the semi-reactive behavior planner introduced by Werling et al. \cite{werling2010optimal}, as it can be tightly integrated into high-level actions and perform reactive obstacle avoidance. While this behavior planner, called Frenet planner in our work, cannot relieve the high-level actions due to its short optimization horizon, it can generate safe and structured behaviors for each high-level action. Frenet planner uses the Frenet frame to model a behavior planning problem. As shown in Figure \ref{fig3:frenet_frame}, the Frenet frame is determined by the tangential vector $\vec{t}_r$ and normal vector $\vec{n}_r$ at a certain point on a reference path. The reference path indicates the ideal path to be followed, such as a centerline of the current or target lane. The cartesian coordinate $\vec{x}=(x,y)$ and the Frenet coordinate $(s,d)$ have the following relation:
\begin{equation} \label{q_function}
\vec{x}(s(t),d(t))=\vec{r}(s(t))+d(s(t))\vec{n}_r(s(t)),
\end{equation}
where $\vec{r}$ is the root point along the reference path, and $s$ and $d$ are the covered arc length and the perpendicular offset, respectively.

The Frenet planner builds on the fact that in a one-dimensional problem with a start state $P_0=[p_0, \dot{p}_0, \ddot{p}_0]$ at $t_0$ and a target state $P_1=[p_1, \dot{p}_1, \ddot{p}_1]$ at $t_1=t_0+K$, quintic polynomials are the optimal connection to minimize the cost function as follows:
\begin{equation} \label{q_function}
C=k_jJ_t + k_t g(K) + k_p h(p_1),
\end{equation}
where $J_t=\int_{t_0}^{t_1}\dddot{p}^2(\tau)d\tau$ is the integral of the square of jerk, $g$ and $h$ are arbitrary functions, and $k_j$, $k_t$, and $k_p$ are the hyperparameters for each term. In particular, Frenet planner generates the longitudinal candidate behaviors $B_{\text{lon}}$ and the lateral candidate behaviors $B_{\text{lat}}$, where both are represented as quintic polynomials, and selects the combined behavior that minimizes the weighted cost function $C_{\text{tot}} = k_{\text{lat}}C_{\text{lat}} + k_{\text{lon}}C_{\text{lon}}$ among all combinations $B_{\text{lon}} \times B_{\text{lat}}$. Note that these candidate behaviors have different end states around the target state.

Figure \ref{fig2:overview} shows how IAHRL uses the Frenet planner to imagine safe and structured behaviors for stopping and lane-following high-level actions in a roundabout scenario. We design each low-level policy to imagine behaviors for a different high-level action. To do so, while both low-level policies regard the current state as the start state $S_{0} = S_{\text{current}}$, they have different longitudinal target states $S_{\text{target}}=[s_{\text{target}}, \dot{s}_{\text{target}}, \ddot{s}_{\text{target}}]$ according to the high-level actions. As for the stopping action, the longitudinal target state is defined as $[s_{\text{target}}, \dot{s}_{\text{target}}, \ddot{s}_{\text{target}}] = [s_{\text{stop}}, 0, 0]$ where $s_{\text{stop}}$ is the target stop position. As for the lane-following action, the longitudinal target state is defined as follows:
\begin{align*}
s_{\text{target}}(t) &= s_{\text{lv}}(t) - [D_0 + \tau \dot{s}_{\text{lv}}(t)], \\
\dot{s}_{\text{target}}(t) &= \dot{s}_{\text{lv}}(t) - \tau \ddot{s}_{\text{lv}}(t), \\
\ddot{s}_{\text{target}}(t) &= \ddot{s}_{\text{lv}}(t_1) - \tau \dddot{s}_{\text{lv}}(t) = \ddot{s}_{\text{lv}}(t_1),
\end{align*}
where $s_{\text{lv}}, \dot{s}_{\text{lv}}, \ddot{s}_{\text{lv}},$ and $\dddot{s}_{\text{lv}}$ are the position, velocity, acceleration, and jerk of the leading vehicle along the same lane, respectively, $D_0$ is the default safety distance, and $\tau$ is the constant. This definition is for ensuring temporal safety distance to the leading vehicle. The lateral target state for both high-level actions is defined as $D_{\text{target}}=[d_{\text{target}}, \dot{d}_{\text{target}}, \ddot{d}_{\text{target}}] = [0, 0, 0]$ to encourage an agent to move parallel to the reference path without the lateral offset. Note that the longitudinal and lateral cost functions for the roundabout scenario are formalized as follows:
\begin{align*}
C_{\text{lat}} &= k^{\text{lat}}_jJ_t + k^{\text{lat}}_t K + k^{\text{lat}}_p d^2_1, \\
C_{\text{lon}} &= k^{\text{lon}}_jJ_t + k^{\text{lon}}_t K + k^{\text{lon}}_p [s_1 - s_{\text{target}}]^2,
\end{align*}
where $K=t_1 - t_0$ is the imagination horizon, $d_1$ is the end lateral position, and $s_1$ is the end longitudinal position. These two cost functions penalize behaviors with slow convergence and lateral offset. In our experiments, we utilize the same target states and cost functions for diverse intersection and lane-change scenarios, except that we consider the centerline of the target lane as the reference path for the lane-change scenario. Please refer to Werling et al. \cite{werling2010optimal} for further details.

\begin{figure*}[t]
\begin{center}
\centerline{\includegraphics[width=0.95\textwidth]{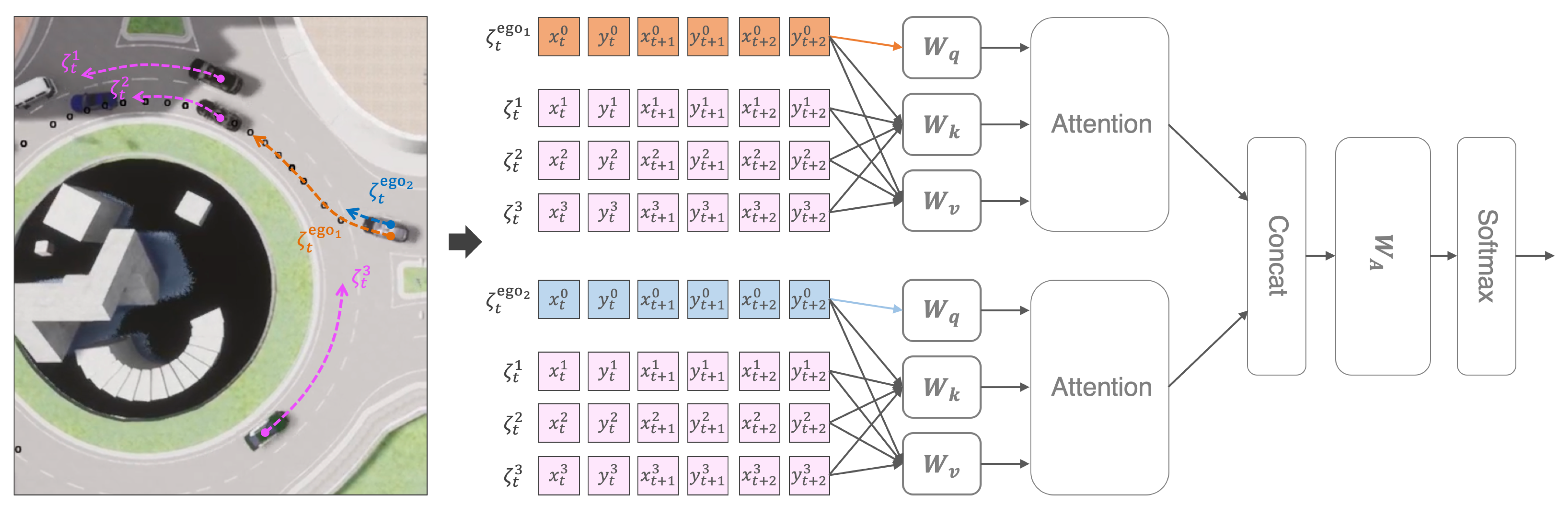}}
\caption{Structure of our attention-based high-level policy. 
While the Key and Value networks, $W_k$ and $W_v$, take as input the imagined behaviors of our agent and surrounding objects, the Query network, $W_q$, only takes as input the imagined behavior of our agent. This proposed structure allows our agent to be permutation-invariant to the order of surrounding objects, while prioritizing our agent over them. In the left image, the route from the current position to a given goal is denoted as a black dotted line. The orange dashed line denotes the agent's behavior of entering the roundabout, and the blue dashed line denotes the agent's behavior of waiting to yield to surrounding vehicles. Similarly, the predicted behaviors of surrounding vehicles are denoted as green dashed lines.}
\label{fig3:high-level_policy}
\end{center}
\vskip -0.1in
\end{figure*}

\subsection{Interaction Inference with Behavior Imagination}
We introduce a new attention mechanism that allows the high-level policy to infer interactions with surrounding objects by interpreting the behaviors imagined with the low-level policies. Our attention mechanism is permutation-invariant to the order of surrounding objects and prioritizes our agent over them, which leads to stable and robust behaviors in dense traffic scenarios. This contrasts with most previous works that naively train an agent with the stack of surrounding object features, which are susceptible to their order.

Before discussing our attention mechanism, we describe the overall structure of our high-level policy shown in Figure \ref{fig3:high-level_policy}. The imagined behavior of a surrounding object at time-step $t$ is a sequence of positions along the X and Y axes and is denoted as $\zeta_t^i = \{(x^i_{t}, y^i_{t}), (x^i_{t+1}, y^i_{t+1}), \dots, (x^i_{t+K}, y^i_{t+K})\}$, where $i$ indicates an index of objects and $K$ is the imagination horizon. Similarly, we denote the imagined behaviors of our agent as $\zeta^{ego_j}_t$, where $j$ indicates an index of behaviors. The behaviors of our agent are imagined by the low-level policies, and the behaviors of surrounding objects are imagined by a given perception module. Note that IAHRL is compatible with any behavior planner and perception module.

The input of our high-level policy is the state $s_t$ defined as an aggregation of the imagined behaviors of our agent and surrounding objects
\begin{equation} \label{eq:state}
s_t = \{\zeta_t^{ego_1},\: \zeta_t^{ego_2}, \: \dots , \: \zeta_t^{ego_n}, \: \zeta_t^1, \: \zeta_t^2, \: \dots , \: \zeta_t^m\},
\end{equation}
where $n$ is the number of our agent’s imagined behaviors, and $m$ is the number of detected surrounding objects. Our high-level policy includes independent attention-based networks for each imagined behavior of our agent. These networks take as input one of our agent's imagined behaviors and all the imagined behaviors of the surrounding objects
\begin{equation} \label{eq:attentino_input}
s^i_t = \{\zeta^{ego_i}, \: \zeta^1_t, \: \zeta^2_t, \: \dots, \: \zeta^m_t\}. 
\end{equation} 
The output of each attention-based network is an embedding vector $m_t^i$ that encodes the interactions induced when our agent follows the imagined behavior $\zeta_t^i$. All the embedding vectors are concatenated as $m_t = \{m_t^1, m_t^2, \dots, m_t^n\}$ and then fed to a fully connected layer $W_A$ that outputs the categorical distribution over the agent’s behaviors. 

To allow our attention mechanism to be permutation-invariant to the order of surrounding objects and prioritize our agent over them, we define the Query matrix $\mathcal{Q}$ as a function that takes our agent’s behavior, $\zeta^{\text{ego}_i}_t$, and random seed vectors, $\eta$, as input. This is inspired by Set Transformer \cite{lee2019set} that is permutation-invariant to the order of input features by using the fixed Query matrix. However, the fixed Query matrix cannot prioritize some input features over others. Our attention mechanism can be written as follows:
\begin{equation} \label{eq:attention}
m^i_t = \sigma(\mathcal{Q}(\zeta_t^{\text{ego}_i}, \eta) \: \mathcal{K}(s_t^i)^T) \: \mathcal{V}(s_t^i) \,.
\end{equation}
In Equation (\ref{eq:attention}), the Query matrix $\mathcal{Q} \in \mathcal{R}^{N_q \times d_q}$ is
\begin{equation} \label{eq:query}
\mathcal{Q}\left(\zeta_t^{\text{ego}_i}, \eta\right)=
\begin{bmatrix}
f_q\left(\zeta_{t}^{ego_i}\right) \\ 
f_q\left(\eta^1\right)\\ 
f_q\left(\eta^2\right)\\
\cdots \\
f_q\left(\eta^{N_q-1}\right) \\
\end{bmatrix} W_q,
\end{equation}
and the Key and Value matrices, $\mathcal{K} \in \mathcal{R}^{(m+1) \times d_q}$ and $\mathcal{V} \in \mathcal{R}^{(m+1) \times d_v}$, are
\begin{equation} \label{eq:attention_matrices}
\! \! \! \mathcal{K}\left(s_t^i\right)=\begin{bmatrix}
f_k\left(\zeta_{t}^{ego_i}\right) \\ 
f_k\left(\zeta_t^1\right)\\
f_k\left(\zeta_t^2\right)\\
\cdots \\
f_k\left(\zeta_t^m\right) \\
\end{bmatrix}W_k, \:
\mathcal{V}\left(s_t^i\right)=\begin{bmatrix}
f_v\left(\zeta_{t}^{ego_i}\right) \\ 
f_v\left(\zeta_t^1\right)\\
f_v\left(\zeta_t^2\right)\\
\cdots \\
f_v\left(\zeta_t^m\right) \\
\end{bmatrix}W_v, \\ [5pt]
\end{equation}
where $W_q, W_k,$ and $W_v$ are projection networks, and $f_q$, $f_k$, and $f_v$ are embedding functions for each matrix. In our experiments, we empirically observed that identity embedding functions are enough to infer interactions from imagined behaviors. Permuting surrounding objects $\{\zeta^{1}_t, \zeta^{2}_t, \dots, \zeta^{m}_t\}$ cannot affect the Query matrix, as it does not depend on them. In contrast, permuting surrounding objects is equivalent to permuting the rows in the Key and Value matrices, as both matrices are linear transformations of $s_t^i$.

Here we offer an intuitive example to demonstrate that our attention mechanism is permutation-invariant to the order of surrounding objects and prioritizes our agent over them. For simplicity, we suppose $N_q=3, m=2, d_q=d_v=1, $ so that $\mathcal{Q} \in \mathcal{R}^{3 \times 1}, \mathcal{K} \in \mathcal{R}^{3 \times 1}, \mathcal{V} \in \mathcal{R}^{3 \times 1}$. Assume our attention mechanism is not permutation-invariant to the order of surrounding objects. It means that the input $s^{i}_t = \{\zeta_t^{\text{ego}_i}, \zeta_t^1, \zeta_t^2\}$ and its permutation of surrounding objects $\hat{s}^{i}_t = \{\zeta_t^{\text{ego}_i}, \zeta_t^2, \zeta_t^1\}$ should result in different outputs $m_t^i$ and $\hat{m}_t^i$. These outputs are calculated as follows:
\begin{align} \label{eq:permutation_invariant_1}
m^i_t &= \sigma 
\begin{pmatrix}
\begin{pmatrix}
q_{i} \\
q_{\eta^1} \\
q_{\eta^2}
\end{pmatrix}
\begin{pmatrix}
k_{i} \!\! & \!\! k_1 \!\! & \!\! k_2
\end{pmatrix}
\end{pmatrix}
\begin{pmatrix}
v_{i} \\ 
v_1 \\
v_2
\end{pmatrix} \nonumber \\[5pt]
&= \sigma 
\begin{pmatrix}
\begin{pmatrix}
q_{i} k_{i} & q_{i} k_1 & q_{i} k_2 \\
q_{\eta^1} k_{i} & q_{\eta^1} k_1 & q_{\eta^1} k_2 \\
q_{\eta^2} k_{i} & q_{\eta^2} k_1 & q_{\eta^2} k_2
\end{pmatrix}
\end{pmatrix}
\begin{pmatrix}
v_{i} \\ 
v_1 \\
v_2
\end{pmatrix} \nonumber \\[5pt]
&= 
\: \: \: \begin{pmatrix}
\sigma (q_{i} k_{i})v_i + \: \sigma (q_{i} k_1)v_1 \: + \sigma (q_{i} k_2)v_2 \\
\sigma (q_{\eta^1} k_{i})v_i + \sigma (q_{\eta^1} k_1)v_1 + \sigma (q_{\eta^1} k_2)v_2 \\
\sigma (q_{\eta^2} k_{i})v_i + \sigma (q_{\eta^2} k_1)v_1 + \sigma (q_{\eta^2} k_2)v_2
\end{pmatrix}, 
\end{align}
\begin{align} \label{eq:permutation_invariant_2}
\hat{m}^i_t &=
\sigma 
\begin{pmatrix}
\begin{pmatrix}
q_{i} \\
q_{\eta^1} \\
q_{\eta^2}
\end{pmatrix}
\begin{pmatrix}
k_{i} \!\! & \!\! k_2 \!\! & \!\! k_1
\end{pmatrix}
\end{pmatrix}
\begin{pmatrix}
v_{i} \\ 
v_2 \\
v_1
\end{pmatrix} \nonumber \\[5pt]
&= \sigma 
\begin{pmatrix}
\begin{pmatrix}
q_{i} k_{i} & q_{i} k_2 & q_{i} k_1 \\
q_{\eta^1} k_{i} & q_{\eta^1} k_2 & q_{\eta^1} k_1 \\
q_{\eta^2} k_{i} & q_{\eta^2} k_2 & q_{\eta^2} k_1
\end{pmatrix}
\end{pmatrix}
\begin{pmatrix}
v_{i} \\ 
v_2 \\
v_1
\end{pmatrix} \nonumber \\[5pt]
&= 
\: \: \: \begin{pmatrix}
\sigma (q_{i} k_{i})v_i + \: \sigma (q_{i} k_2)v_2 \: + \sigma (q_{i} k_1)v_1 \\
\sigma (q_{\eta^1} k_{i})v_i + \sigma (q_{\eta^1} k_2)v_2 + \sigma (q_{\eta^1} k_1)v_1 \\
\sigma (q_{\eta^2} k_{i})v_i + \sigma (q_{\eta^2} k_2)v_2 + \sigma (q_{\eta^2} k_1)v_1
\end{pmatrix},
\end{align}
where $q_{\eta^1}$ and $q_{\eta^2}$ are independent of the input $s_t^i$. Equations (\ref{eq:permutation_invariant_1}) and (\ref{eq:permutation_invariant_2}) show that permuting the surrounding objects only affects the order of the summation in each row, and both outputs, $m_t^i$ and $\hat{m}_t^i$, are identical. This result is in contradiction to the assumption, and thus our attention mechanism must be permutation-invariant to the order of surrounding objects. 

Equations (\ref{eq:permutation_invariant_1}) and (\ref{eq:permutation_invariant_2}) also illustrate that permuting an entire state $s^i_t$, including behaviors of our agent and surrounding objects, will change the output of our attention mechanism, as the value of the Query matrix is a linear transformation of $\zeta^{\text{ego}_i}_t$. In other words, our attention mechanism is trained to prioritize our agent over surrounding objects by utilizing the positional information of our agent.

\subsection{High-Level Policy Update with Imagined Behaviors}
The high-level policy should be trained to select the low-level policy that imagines the most interactive behavior. Therefore, IAHRL uses the sum of discounted low-level rewards obtained by following the activated low-level policy within H time steps, $\sum_{t'=t}^{t+H-1}\gamma^{t'-t}r(s_{t'},a_{t'})$, as the high-level reward, $\bar{r}(s_t,z_t)$. This is distinct from pre-training the high-level policy and then naively combining it with the low-level policies. While IAHRL is compatible with any RL algorithm, we use Soft Actor-Critic (SAC) \cite{haarnoja2018soft}, which is a state-of-the-art off-policy RL algorithm providing robust and sample-efficient learning. We first maximize the maximum entropy (MaxEnt) RL objective that augments the expected sum of rewards with the expected entropy of the policy as follows:
\begin{equation} \label{eq:objective}
J(\pi) = \sum_{t=0}^T \mathbb{E}_{(s_t,z_t) \sim \rho_\pi} [\bar{r}(s_t,z_t) + \alpha \mathcal{H}(\pi_\phi(\cdot|s_t))],
\end{equation}
where $\phi$ is the parameter of the high-level policy, $\rho_\pi$ is the state-action marginals of the distribution induced by the high-level policy, $\alpha$ is the temperature parameter, and $\mathcal{H}(\pi_\phi(\cdot|s_t))$ is the entropy of the high-level policy. Note that we consider discrete action space in the following paragraphs, as our high-level policy selects behaviors imagined by the low-level policies, not continuous atom actions.

To maximize the objective in Equation (\ref{eq:objective}), we conduct soft policy iteration that alternates between policy evaluation and policy improvement in the MaxEnt RL framework. The policy evaluation process involves learning soft high-level state-action value function $Q^\pi(s_t,z_t)$, and its objective can be written as follows:
\begin{equation} \label{eq:q_objective}
\begin{aligned}
& \! \! \! J_{Q^\pi}(\theta) = \mathbb{E}_{(s_t,z_t) \sim D}\Big[\frac{1}{2}\big(Q^\pi_\theta(s_t,z_t) \\
& \: \: \: \: \qquad - (\bar{r}(s_t,z_t) + \gamma \mathbb{E}_{s_{t+H} \sim p(s_t,z_t)}[V_{\bar{\theta}}^\pi(s_{t+H})])\big)^2\Big], \\[4pt]
\end{aligned}
\end{equation}
with the soft high-level state value function
\begin{equation} \label{eq:value_function}
V_{\bar{\theta}}^\pi(s_t) = \sum_{z_t \in Z} \pi_\phi(z_t|s_t)(Q^\pi_{\bar{\theta}}(s_t,z_t) - \log \pi_\phi(z_t|s_t)),
\end{equation}
where $\theta$ and $\bar{\theta}$ are the parameters of the soft high-level state-action value function and the target function of the value function, respectively, and $D$ is a replay buffer. Leveraging the target value function $Q^\pi_{\bar{\theta}}(s_t,z_t)$, where $\bar{\theta}$ is updated as an exponential moving average of $\theta$, can improve the stability of our training \cite{mnih2015human, lillicrap2015continuous}. While the soft state value function can be represented as a separate model, we empirically observed that estimating this function based on the policy and the soft state-action value function is sufficient to learn urban driving tasks in our experiments. 

In the policy improvement process, we update our high-level policy towards the exponential of the soft high-level state-action value function obtained from the policy evaluation process. This update ensures the improvement of the policy with regard to its soft value and can be formalized as follows: 
\begin{equation} \label{eq:policy_objective}
J_\pi(\phi) = \mathbb{E}_{s_t \sim D}[D_{KL}(\pi_\phi(z_t|s_t) \| \frac{\exp(Q^\pi_\theta(s_t,z_t))}{Z_\theta(s_t)})]
\end{equation}
where $D_{KL}(\cdot \| \cdot)$ is the KL-divergence, and $Z_\theta(s_t)$ is the partition function that normalizes the distribution. The partition function is intractable, but we can ignore this term as it is independent of $\phi$. Then, the objective in Equation (\ref{eq:policy_objective}) can be simplified as follows:
\begin{equation} \label{eq:policy_objective_simplified}
J_\pi(\phi) = \mathbb{E}_{s_t \sim D}[\sum_{z_t \in Z} \pi_\phi(z_t|s_t)(\log \pi_\phi(z_t|s_t) - Q^\pi_\theta(s_t,z_t))].
\end{equation}

\begin{figure*}[t]
\vskip 0.05in
\begin{center}
\centerline{\includegraphics[width=0.98\textwidth]{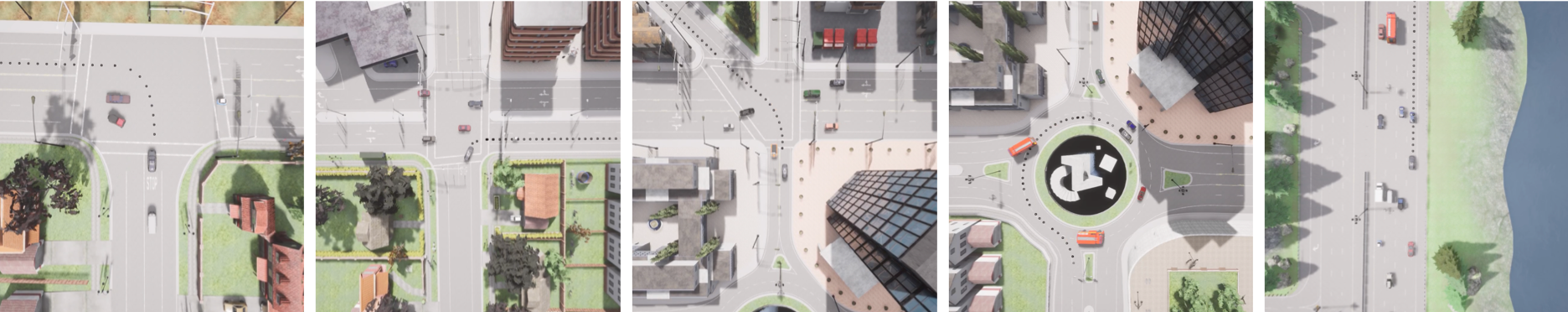}}
\caption{Urban driving tasks introduced in our work. The route to a given goal is represented as a black dotted line. All spawned vehicles are set to ignore traffic signals, so our agent should consider interactions with surrounding vehicles to solve these tasks. Unlike other tasks, spawned vehicles in the lane-change task are initialized with different speeds from 15km/h to 30km/h.}
\label{environments}
\end{center}
\vskip -0.10in
\end{figure*}

The temperature parameter, $\alpha$, in Equation (\ref{eq:objective}) controls the trade-off between exploration and exploitation. This parameter should be carefully tuned, as it depends heavily on a task, and sub-optimal temperature can lead to significant performance degradation. To address this issue, an extension of SAC \cite{haarnoja2018soft-extension} introduces an automatic temperature tuning method that adjusts the temperature so that the entropy of the policy matches a constant target entropy. The objective for the temperature parameter is formalized as follows:
\begin{equation} \label{eq:alpha_objective}
\! \! \! J(\alpha) = \mathbb{E}_{s_t \sim D}[\sum_{z_t \in Z} \pi_\phi(z_t|s_t)(-\alpha(\log \pi_\phi(z_t|s_t) + \bar{\mathcal{H}}))].
\end{equation}
where $\bar{\mathcal{H}}$ is the target entropy. The overall training procedure of our algorithm is described in Algorithm \ref{alg:training}.

\begin{algorithm}[t]
   \caption{Overall Training Procedure of IAHRL}
   \label{alg:training}
\algsetup{linenosize=\small}
\begin{algorithmic}[1]
   \STATE {\bfseries Given:} Low-level policies $\pi_{z}(a|s)$
   \vspace{0.05mm}
   \STATE Initialize parameter vectors $\theta, \bar{\theta}, \phi$, and replay buffer $D$
   \vspace{0.05mm}
   \FOR{each iteration}
   \vspace{0.05mm}
   \FOR{$t=0,\, H,\, \dots, T-1$}
   \vspace{0.05mm}
   \STATE Select high-level action $z_t \sim \pi_\phi(z_t|s_t)$
   \vspace{0.05mm}
   \FOR{$t'=t,\, t+1,\, \dots, t+H-1$}
   \vspace{0.05mm}
   \STATE Execute action $a_{t'} \sim \pi_{z_t}(a_{t'}|s_{t'})$
   \vspace{0.05mm}
   \STATE Observe next state $s_{t'+1}$ and reward $r_{t'}$
   \vspace{0.05mm}
   \ENDFOR
   \vspace{0.05mm}
   \STATE Calculate high-level reward $\bar{r}_t = \sum_{t'=t}^{t+H-1}\gamma^{t'-t}r_{t'}$ \\
   \vspace{0.05mm}
   \STATE Add transition to replay buffer \\
          $D \leftarrow D \cup \{(s_t, z_t, \bar{r}_t, s_{t+H})\}$
   \vspace{0.05mm}
   \FOR{$l=1,\, 2,\, \dots, L$}
   \vspace{0.1mm}
   \STATE Sample random mini-batch of $N_b$ transitions \\
          $\{(s_t, z_t, \bar{r}_t, s_{t+H})\}_{N_b}$ from replay buffer $D$
   \vspace{0.5mm}
   \STATE Update $\pi_\phi(z_t|s_t)$ with Equation (\ref{eq:policy_objective_simplified}) \\[2pt]
          $\phi \leftarrow \phi - \delta \, \nabla_\phi J_\pi(\phi)$
   \vspace{0.5mm}
   \STATE Update $Q_\theta^\pi(s_t,z_t)$ with Equation (\ref{eq:q_objective}) \\[2pt]
          $\theta \leftarrow \theta - \delta \, \nabla_\theta J_{Q^\pi}(\theta)$
   \vspace{0.5mm}
   \STATE Update $\alpha$ with Equation (\ref{eq:alpha_objective}) \\[2pt]
          $\alpha \leftarrow \alpha - \delta \, \nabla_\alpha J(\alpha)$
   \vspace{0.05mm}   
   \STATE Update $Q_{\bar{\theta}}^\pi(s_t,z_t)$ with moving average \\[2pt]
          $\bar{\theta} \leftarrow \omega \theta - (1-\omega) \bar{\theta}$
   \vspace{0.05mm}      
   \ENDFOR
   \ENDFOR
   \ENDFOR
   \vspace{0.05mm}
\end{algorithmic}
\end{algorithm}

\section{Experiments}
Our experiments aim to answer the following questions: 1) Can IAHRL enable an agent to perform safe and interactive behaviors in real-world navigation tasks?; 2) Can IAHRL properly infer interactions with surrounding objects from imagined behaviors?; 3) Can IAHRL ensure robust performance against imagination error?; and 4) How does the imagination horizon, $K$, affect the performance of IAHRL? To answer these questions, we introduce five urban driving tasks using the autonomous driving simulator CARLA \cite{dosovitskiy2017carla}. An agent in these tasks should follow traffic rules and perform safe and interactive behaviors to reach a given goal.

\subsection{Urban Driving Tasks}
Urban driving tasks are among the most challenging real-world navigation tasks. We introduce the five complex urban driving tasks containing three intersection tasks, a roundabout task, and a lane-change task. In contrast to typical settings, all spawned vehicles in these tasks are initialized to ignore any traffic signals. This setting makes it more difficult for our agent to solve these tasks without interactions with surrounding vehicles. Note that we use simple reward functions for these tasks, as implementing low-level policies with the optimization-based behavior planner greatly reduces the burden of manually specifying reward functions for safety-aware behaviors following traffic rules, such as keeping a safe distance, or not crossing the center line. Figure \ref{environments} shows the urban driving tasks introduced in our work, and further details about them are described in the following paragraphs.

\textbf{Three-/four-/five-way intersections:} We introduce three intersection tasks with different numbers of involved road segments. These tasks require an agent to reach a given goal point as fast as possible without collisions. Seven vehicles are spawned at random locations within 70 meters from the center of the intersection. These vehicles are initialized to start in a stationary state. The behavior planner used in these tasks generates two distinct behaviors ($n=2$): one with a target speed above 5km/h and one below 5km/h. The reward function is designed as follows:
\begin{equation} \label{eq:intersection_reward}
r(s_t, a_t) = \lambda_v \, \frac{v_t}{v_{max}} - \lambda_c\mathbbm{1}_c(s_t,a_t) - \lambda_s\mathbbm{1}_s(s_t,a_t),
\end{equation}
where $v_t$ is a current speed, $v_{max}$ is a maximum speed, $\mathbbm{1}_c$ is an indicator function of whether our agent collides or fails to keep a safe distance, $\mathbbm{1}_s$ is an indicator function of whether our agent survives, and $\lambda_v$, $\lambda_c$, and $\lambda_s$ are hyperparameters to balance each of these terms, respectively. Note that, while these interaction tasks share the same setting, each task has its own unique traffic patterns.

\textbf{Roundabout:} We introduce a roundabout task that has four involved road segments. This task has the same settings as the intersection tasks except that 10 vehicles are spawned within 80 meters from the center of the roundabout. Note that the behavior patterns of vehicles at the roundabout are significantly different from those observed at typical intersections.

\textbf{Lane-change:} We introduce a lane-change task that requires an agent to reach a given target lane as fast as possible without awkward deceleration or collisions. To simulate high-traffic scenarios, we spawn 20 vehicles within a 50-meter radius of our agent. Unlike other tasks, all spawned surrounding vehicles and our agent are initialized with different speeds from 10km/h to 30km/h. 
The behavior planner used in the lane-change task generates three distinct behaviors ($n=3$): one that reaches the target lane and the others that follow the current lane with a target speed above or below 5 km/h. The reward function is designed as follows:
\begin{equation} \label{eq:lanechange_reward}
r(s_t, a_t) = \lambda_v \frac{v_t - v_{des}}{v_{max}} - \lambda_c\mathbbm{1}_c(s_t,a_t) - \lambda_s\mathbbm{1}_s(s_t,a_t),
\end{equation}
where $v_{des}$ is a desired speed for lane keeping.

\begin{table}[t]
\caption{Hyperparameters \label{tab:hyperparameters}}
\begin{center}
\resizebox{0.95\columnwidth}{!}{
\begin{tabular}{c c}
\toprule
\qquad HYPERPARAMETER \qquad & \qquad VALUE \qquad \\
\midrule
\qquad Episode Horizon ($T$) \qquad & \qquad 600 \qquad \\
\vspace{0.2mm}
\qquad High-Level Action Horizon ($H$) \qquad & \qquad 30 \qquad \\
\vspace{0.2mm}
\qquad Imagination Horizon ($K$) \qquad & \qquad 5 \qquad \\
\vspace{0.2mm}
\qquad Batch Size ($N_b$) \qquad & \qquad 128 \qquad \\
\vspace{0.2mm}
\qquad Replay Buffer Size ($|D|$) \qquad & \qquad 50000 \qquad \\
\vspace{0.2mm}
\qquad Learning Rate ($\delta$) \qquad & \qquad 0.00003 \qquad \\
\vspace{0.2mm}
\qquad Discount Factor ($\gamma$) \qquad & \qquad 0.99 \qquad \\
\vspace{0.2mm}
\qquad $\text{Adam}_1$ ($\beta_1$) \qquad & \qquad 0.9 \qquad \\
\vspace{0.2mm}
\qquad $\text{Adam}_2$ ($\beta_2$) \qquad & \qquad 0.999 \qquad \\
\vspace{0.2mm}
\qquad Temperature ($\alpha$) \qquad & \qquad 0.4 \qquad \\
\vspace{0.2mm}
\qquad The Number of Detection Vehicles ($m$) \qquad & \qquad 5 \qquad \\
\vspace{0.2mm}
\qquad The Number of Rows of $\mathcal{Q}$ ($N_q$) \qquad & \qquad 6 \qquad \\
\vspace{0.2mm}
\qquad The Number of Columns of $\mathcal{Q}$ ($d_q$) \qquad & \qquad 24 \qquad \\
\vspace{0.2mm}
\qquad The Number of Columns of $\mathcal{V}$ ($d_v$) \qquad & \qquad 24 \qquad \\
\vspace{0.2mm}
\qquad Query Enbedding Function ($f_q$) \qquad & \qquad $I$ \qquad \\
\vspace{0.2mm}
\qquad Key Enbedding Function ($f_k$) \qquad & \qquad $I$ \qquad \\
\vspace{0.2mm}
\qquad Value Enbedding Function ($f_v$) \qquad & \qquad $I$ \qquad \\
\vspace{0.2mm}
\qquad Safety Distance ($D_0$) \qquad & \qquad 3 \qquad \\
\vspace{0.2mm}
\qquad Safety Time ($\tau$) \qquad & \qquad 2.5 \qquad \\
\vspace{0.2mm}
\qquad Collision Weight ($\lambda_c$) \qquad & \qquad 2 \qquad \\
\vspace{0.2mm}
\qquad Survive Weight ($\lambda_s$) \qquad & \qquad 1 \qquad \\
\vspace{0.2mm}
\qquad Velocity Weight ($\lambda_v$) (Lane-Change) \qquad & \qquad 2 \qquad \\
\vspace{0.2mm}
\qquad Velocity Weight ($\lambda_v$) (Others) \qquad & \qquad 1 \qquad \\
\vspace{0.2mm}
\qquad The Number of Updates ($L$) \qquad & \qquad 1 \qquad \\
\vspace{0.2mm}
\qquad Target Smoothing Coefficient ($\omega$) \qquad & \qquad 0.005 \qquad \\
\bottomrule
\end{tabular}
}
\end{center}
\end{table}

\subsection{Baselines}
We compare IAHRL against four baselines: 1) a flat agent that randomly selects the optimization-based planner (RANDOM), 2) a hierarchical agent that randomly selects the optimization-based planner every H step (H-RANDOM), 3) an open-source autonomous driving agent provided by CARLA (CARLA) \cite{dosovitskiy2017carla}, and 4) the most closely related recent work that trains a hierarchical agent with an optimization-based behavior planner (H-CtRL) \cite{li2021safe}. While IAHRL infers interactions from the predicted behaviors by integrating imagination into HRL, H-CtRL infers interactions from the current state, including positions, speeds, and yaw angles of surrounding vehicles. To make a fair comparison, the same behavior planner and controller are used in our agent and baselines. We implemented the behavior planner as the Frenet planner \cite{werling2010optimal} and used PID controllers to get atom actions, including throttle, brake, and SWA.

\subsection{Implementation Details}
IAHRL trains two learnable models: the high-level policy $\pi_\phi(z|s)$ and the high-level state-action value function $Q_\theta^\pi(s,z)$. Both models are implemented with neural networks and have the same attention-based structure, except that the policy includes an additional softmax layer to output a categorical distribution over behaviors. In this attention-based structure, the Query, Key, and Value projection functions, $W_q, W_k, $ and $W_v$, respectively, are represented as neural networks having two hidden layers with ReLU activations. The parameters of all models are updated with the Adam optimizer. Although learning navigation strategies for urban autonomous driving is a challenging problem, we empirically observed that IAHRL quickly reaches asymptotic performance within about 150K environmental steps. We believe that this sample efficiency is attributed to our hierarchical decomposition based on safe and structured behaviors. Table \ref{tab:hyperparameters} describes the hyper-parameters used in our experiment, and they were tuned with the coarse grid search. All experiments were run on a PC with a 3.20 GHz Intel i9-12900KF Processor, a GeForce RTX 2080 Ti GPU, and 64GB of RAM.

\begin{table*}[!t]
\vskip 0.2in
\caption{Quantitative Results on Intersection Tasks \label{tab:numerical_results_inter}}
\begin{center}
\resizebox{1.0\textwidth}{!}{
\begin{tabular}{c cccc | cccc | cccc}
\toprule
& \multicolumn{4}{c}{THREE-WAY} & \multicolumn{4}{c}{FOUR-WAY} & \multicolumn{4}{c}{FIVE-WAY}\\
\toprule
& \: AR ($\bar{r}$) $\uparrow$ & AS ($t$) $\downarrow$ & SR ($\%$) $\uparrow$ & CR ($\%$) $\downarrow$ \qquad 
& \: AR ($\bar{r}$) $\uparrow$ & AS ($t$) $\downarrow$ & SR ($\%$) $\uparrow$ & CR ($\%$) $\downarrow$ \qquad 
& \: AR ($\bar{r}$) $\uparrow$ & AS ($t$) $\downarrow$ & SR ($\%$) $\uparrow$ & CR ($\%$) $\downarrow$\\
\midrule
RANDOM &  -133.3 & 914.3 & 0.57 & 0.01  \qquad & -180.5 & 887.8 & 0.65 & 0.06 \qquad &  \: -86.9 & 918.8 & 0.54 & 0.03\\
H-RANDOM & \: -93.6 & 411.3 & 0.85 & 0.15  \qquad & -152.5 & 445.8 & 0.79 & 0.21 \qquad & -101.3 & 445.8 & 0.83 & 0.17\\
CARLA & -237.9 & 339.6 & 0.84 & 0.16  \qquad & -692.6 & 532.8 & 0.41 & 0.48 \qquad & -505.2 & 470.1 & 0.57 & 0.31\\
H-CtRL & -166.8 & 320.1 & 0.85 & 0.15  \qquad & \: -86.8 & 396.5 & 0.89 & 0.10 \qquad & \: \: \ 8.1 & 373.5 & 0.95 & 0.04\\
\textbf{OURS} & \: \ \textbf{63.1} & \textbf{253.2} & \textbf{0.96} & \textbf{0.03}  \qquad & \: \ \textbf{30.5} & \textbf{291.9} & \textbf{0.95} & \textbf{0.05} \qquad & \: \ \textbf{79.2} & \textbf{303.3} & \textbf{0.97} & \textbf{0.03}\\
\bottomrule
\end{tabular}
}
\end{center}
\end{table*}

\begin{table*}[!t]
\caption{Quantitative Results on Roundabout and Lane-change Tasks \label{tab:numerical_results_others}}
\begin{center}
\resizebox{0.72\textwidth}{!}{
\begin{tabular}{c cccc | cccc}
\toprule
& \multicolumn{4}{c}{ROUNDABOUT} & \multicolumn{4}{c}{LANE-CHANGE}\\
\toprule
& \: AR ($\bar{r}$) $\uparrow$ & AS ($t$) $\downarrow$ & SR ($\%$) $\uparrow$ & CR ($\%$) $\downarrow$ \qquad 
& \: AR ($\bar{r}$) $\uparrow$ & AS ($t$) $\downarrow$ & SR ($\%$) $\uparrow$ & CR ($\%$) $\downarrow$\\
\midrule
RANDOM & \: -80.7 & 996.7 & 0.02 & 0.05  \qquad & -131.9 & 241.7 & 0.47 & 0.53\\
H-RANDOM & \: \ 21.0 & 821.4 & 0.39 & 0.52  \qquad & \ -56.5 & \: 96.9 & 0.58 & 0.42\\
CARLA & -495.1 & 670.8 & 0.21 & 0.64  \qquad & - & - & - & -\\
H-CtRL & \: \ 86.8 & 749.4 & 0.86 & 0.12  \qquad & \ -31.3 & 112.5 & 0.81 & 0.19\\
\textbf{OURS} \qquad & \: \textbf{175.0} & \textbf{525.9} & \textbf{0.98} & \textbf{0.01}  \qquad & \: \ \textbf{-7.2} & \: \textbf{79.8} & \textbf{1.00} & \textbf{0.00}\\
\bottomrule
\end{tabular}
}
\end{center}
\end{table*}

\subsection{Experimental Results and Analysis}
We use the following essential evaluation metrics: average episode step (AS), average episode return (AR), success rate (SR), and collision rate (CR) \cite{dosovitskiy2017carla, isele2018navigating, anzalone2022end, hwang2022autonomous, cai2022dq, wang2023efficient}. The average episode step captures how quickly an agent reaches a given goal. The average episode return reflects the effectiveness of an agent’s behaviors in terms of rewards. The success rate and the collision rate provide a more intuitive perspective of an agent’s behavior. All these metrics can be defined as follows:
\begin{align*}
\text{AS} &= \frac{1}{E_{\text{tot}}} \sum_{i=1}^{E}T_{\text{end}_i}, \quad &\text{AR} &= \frac{1}{E_{\text{tot}}} \sum_{i=1}^{E}R_i, \\
\text{SR} &= \frac{E_{\text{suc}}}{E_{\text{tot}}} \times 100, \quad &\text{CR} &= \frac{E_{\text{col}}}{E_{\text{tot}}} \times 100,
\end{align*}
where $T_{\text{end}_i}$ is the time step when an agent reaches a given goal within an episode, $R_i$ is the return of an episode, $E_{\text{tot}}$ is the total number of episodes, $E_{\text{suc}}$ is the number of success episodes where success denotes whether an agent reaches a given goal within $T$ time steps without collision, and $E_{\text{col}}$ is the number of collision episodes where collision denotes whether an agent maintains a safe distance during an episode. We computed these metrics over $E_{\text{tot}}=100$  episodes.

\begin{figure}[t]
\vskip 0.05in
\begin{center}
\centerline{\includegraphics[width=1.0\columnwidth]{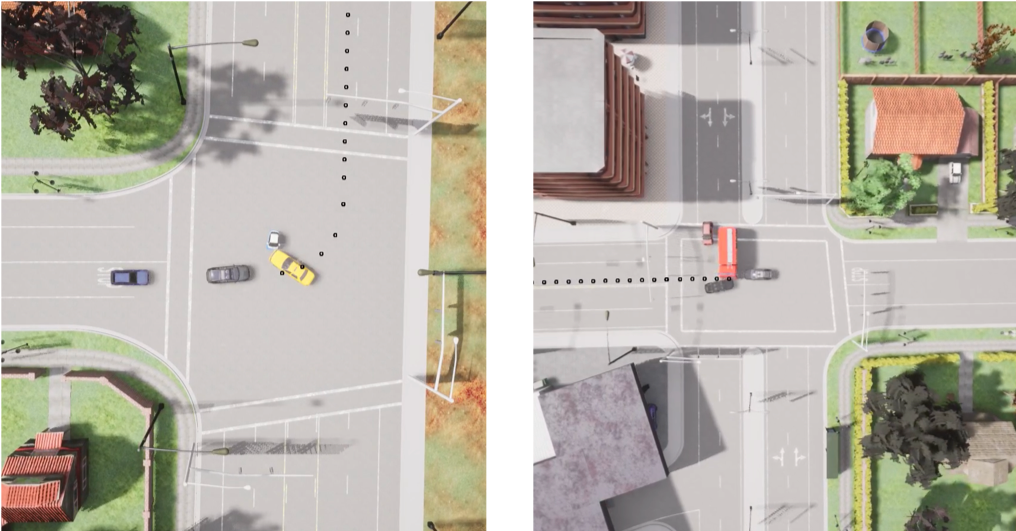}}
\caption{Visualization of failure cases in intersection tasks. The route to the given goal that our agent is supposed to follow is completely blocked due to collisions between surrounding vehicles.}
\label{failure_cases}
\end{center}
\end{figure}

\begin{figure*}[t]
\vskip -0.1in
\begin{center}
\centerline{\includegraphics[width=0.83\textwidth]{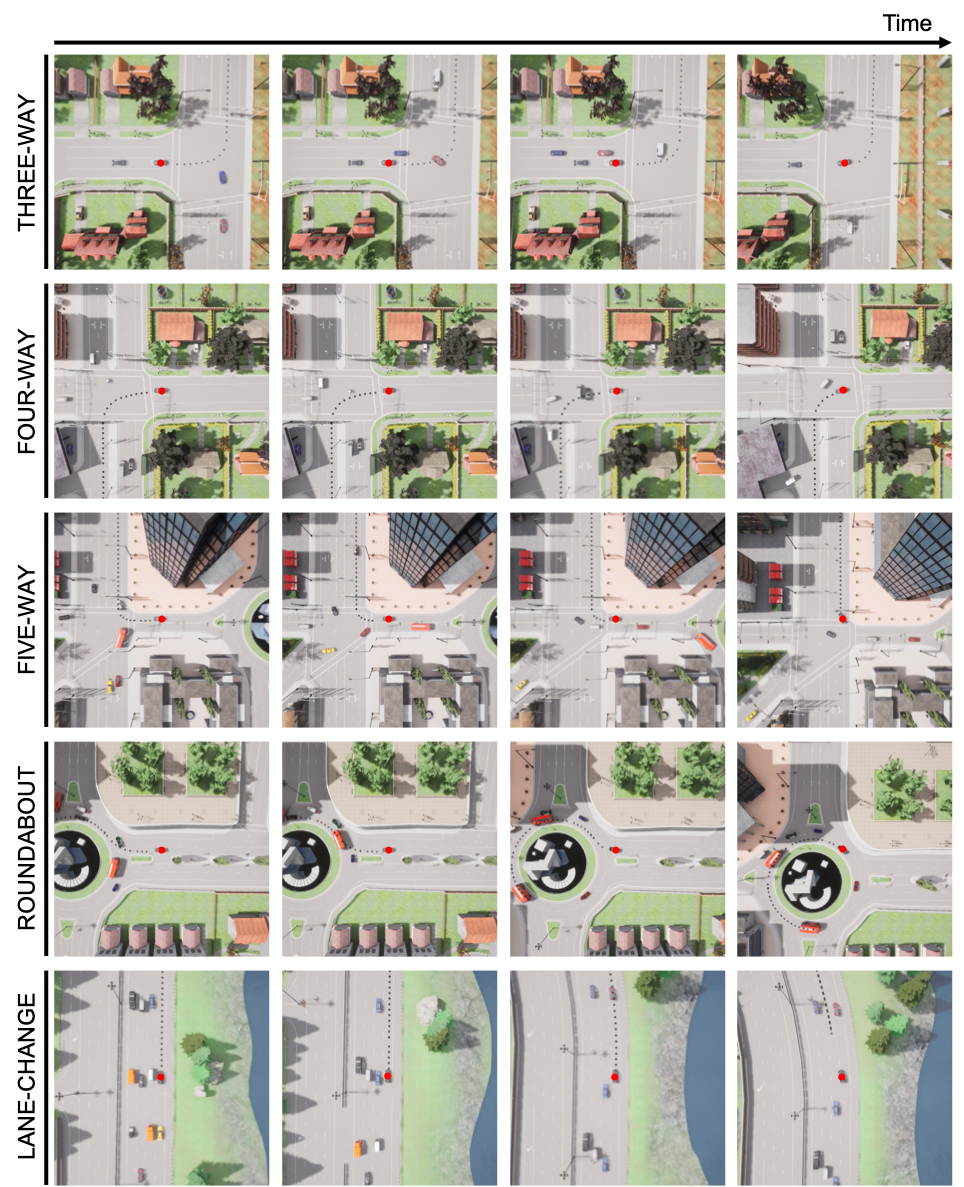}}
\vskip -0.1in
\caption{Visualization of learned behaviors for urban driving tasks. The route to a given goal is represented as a black dotted line. Each row corresponds to a sequence of images that show the learned behaviors of our agent within one specific episode. These qualitative results indicate that our navigation algorithm enables an agent to handle diverse and complex scenarios in urban driving environments.}
\label{visualization}
\end{center}
\vskip -0.1in
\end{figure*}

\textit{\textbf{1) Can IAHRL enable an agent to perform safe and interactive behaviors in real-world navigation tasks?}}
Table \ref{tab:numerical_results_inter} shows the numerical training results on three intersection tasks. We observed that RANDOM drives very slowly and does not encounter other vehicles at intersections, leading to low collision rates and high average episode steps. CARLA obtains a lower average episode steps and a higher success rate than RANDOM in the three-way intersection task. However, its performance drops drastically in the other two intersection tasks, which require an agent to handle more diverse behavior patterns. This indicates that hand-crafted algorithms cannot scale to diverse real-world driving tasks. While H-CtRL attains the best performance among the baselines, our agent solves these tasks much faster than H-CtRL, while maintaining safety. This demonstrates that IAHRL enables an agent to perform safe and interactive behaviors in diverse intersection tasks.

\begin{figure*}[t]
\vskip 0.05in
\begin{center}
\centerline{\includegraphics[width=0.95\textwidth]{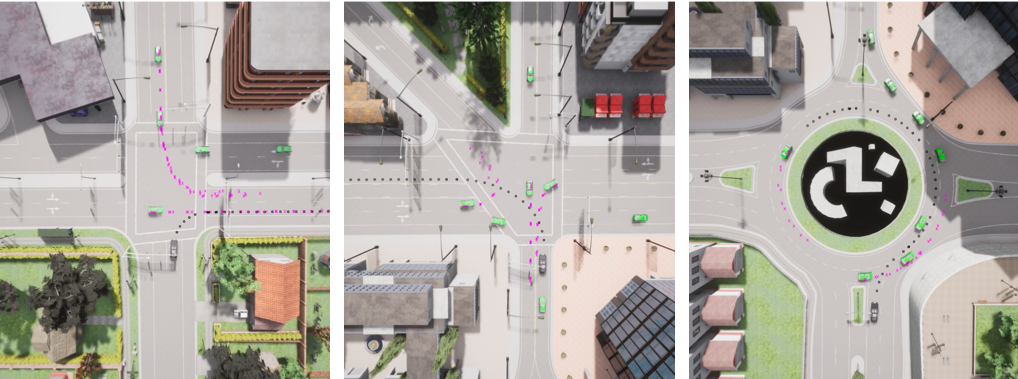}}
\caption{Attention visualization. The route to a given goal is represented as a black dotted line. The top three predicted behaviors of surrounding vehicles that received the most attention are denoted as magenta dotted lines. These qualitative results suggest that our navigation algorithm enables an agent to give attention to surrounding vehicles in a manner akin to human drivers.
}
\label{attention_results}
\end{center}
\vskip -0.1in
\end{figure*}

Table \ref{tab:numerical_results_others} describes the numerical training results on the roundabout and lane-change tasks. RANDOM, H-RANDOM, and CARLA have success rates lower than 50$\%$ in the roundabout task. H-CtRL tends to adopt conservative behaviors, waiting until all vehicles have left the roundabout. In contrast, our agent harmoniously cuts in between vehicles in the roundabout. This enables our agent to achieve a lower average episode step and a higher success rate than all baselines. The lane-change task has a shorter horizon compared to the roundabout task, but the spawned surrounding vehicles in the lane-change task move at higher speeds. This setting makes it more critical to consider interactions with surrounding vehicles when solving the task. Note that our agent solves all evaluation episodes without collisions, significantly outperforming all the baselines. 

As described in Figure \ref{failure_cases}, we empirically confirm that collisions at the intersections and the roundabout are primarily due to collisions between the surrounding vehicles, rather than between our agent and surrounding vehicles. Our agent approaches crashed vehicles at a low speed and stops before them, leaving a gap that is less than the safe distance. These collisions are unavoidable regardless of how well our agent interacts with surrounding vehicles. Although we could potentially improve the simulator's navigation algorithms to reduce collisions, we are leaving that as a future work since it is not relevant to what we focus on here.

Figure \ref{visualization} visualizes the learned behaviors of IAHRL in the introduced urban driving tasks. The first three rows show that our agents perform safe and interactive behaviors at the intersections. Note that these intersections have their own unique traffic patterns due to their different road structures. The fourth row shows our agent safely entering the roundabout without interrupting traffic flows. The fifth row shows the learned behaviors in the lane-change task. We observed that our agent does not perform reckless lane-change behaviors in heavy traffic, but instead performs cooperative lane-change behaviors without causing awkward or dangerous deceleration of surrounding vehicles. These qualitative results imply that IAHRL can enable an agent to handle diverse and complex driving scenarios in urban environments.

\textit{\textbf{2) Can IAHRL properly infer interactions with surrounding objects from imagined behaviors?}}
To thoroughly understand whether IAHRL properly infers interactions with surrounding vehicles, we analyzed the attention weight vector $\sigma(\mathcal{Q}\mathcal{K}^T)$ corresponding to the selected imagined behavior of our agent. This weight vector encodes how much the agent gives attention to each surrounding vehicle as a value between 0 and 1. Figure \ref{attention_results} shows which surrounding vehicles our agent pays attention to. The top three imagined behaviors of surrounding vehicles that received the most attention are highlighted with magenta dotted lines. In the intersection tasks, our agent pays attention not to vehicles close to our agent, but to vehicles whose imagined behaviors are likely to overlap with our agent's behavior. Specifically, at the five-way intersection, our agent focuses on the vehicle making a U-turn in the upper left direction, rather than the vehicles on the left entering the opposite lane or immediately behind. In the roundabout task, the agent focuses on the vehicle in the roundabout that is currently far away but approaching, rather than vehicles that are close but moving away. These attention results are very similar to how humans observe surrounding vehicles while driving, which implies that our agent can properly infer interactions with surrounding vehicles from imagined behaviors.

\begin{table}[!t]
\caption{Performance Comparison based on \\ the Scale of Imagination Errors \label{tab:imagination_errors}}
\begin{center}
\resizebox{0.9\columnwidth}{!}{
\begin{tabular}{c cccc}
\toprule
& \multicolumn{4}{c}{FIVE-WAY}\\
\toprule
& \: AR ($\bar{r}$) $\uparrow$ & AS ($t$) $\downarrow$ & SR ($\%$) $\uparrow$ & CR ($\%$) $\downarrow$ \\
\midrule
$\sigma_{xy}=0.0$ & \: 79.2 & 303.3 & 0.97 & 0.03 \\
$\sigma_{xy}=0.1$ & \: 76.3 & 318.3 & 0.89 & 0.11 \\
$\sigma_{xy}=0.3$ & \: 61.7 & 326.1 & 0.86 & 0.14 \\
$\sigma_{xy}=0.5$ & \: 62.8 & 323.7 & 0.86 & 0.13 \\
\midrule
$\sigma_{xy}=1.0$ & \: 20.5 & 321.0 & 0.78 & 0.22 \\
\bottomrule
\end{tabular}
}
\end{center}
\vskip -0.15in
\end{table}

\textit{\textbf{3) Can IAHRL ensure robust performance against imagination errors?}}
Imagined behaviors of surrounding vehicles often have prediction errors in autonomous driving scenarios. Standard navigation algorithms are susceptible to these imagination errors, as they directly depend on imagined behaviors. In contrast, IAHRL can enable an agent to perform robust behaviors against imagination errors by learning to interpret imagined behaviors. To demonstrate this, we analyzed how the prediction errors of imagined behaviors affect IAHRL. The prediction errors are implemented by adding Gaussian noises with different standard deviations, $\sigma_{xy}$, to imagined behaviors. Table \ref{tab:imagination_errors} illustrates the performance of IAHRL according to the standard deviation scale of Gaussian noise on the five-way intersection task. We observed that although imagination errors lead to slightly worse performance, IAHRL retains competitive performance across a wide range, except for a standard deviation of 1.0. Note that Gaussian noise with a standard deviation of 1.0 can be larger than the width of a lane. This result implies that IAHRL can handle the prediction errors of imagined behaviors by learning to interpret them.

\begin{figure}[t]
\vskip 0.05in
\begin{center}
\centerline{\includegraphics[width=1.0\columnwidth]{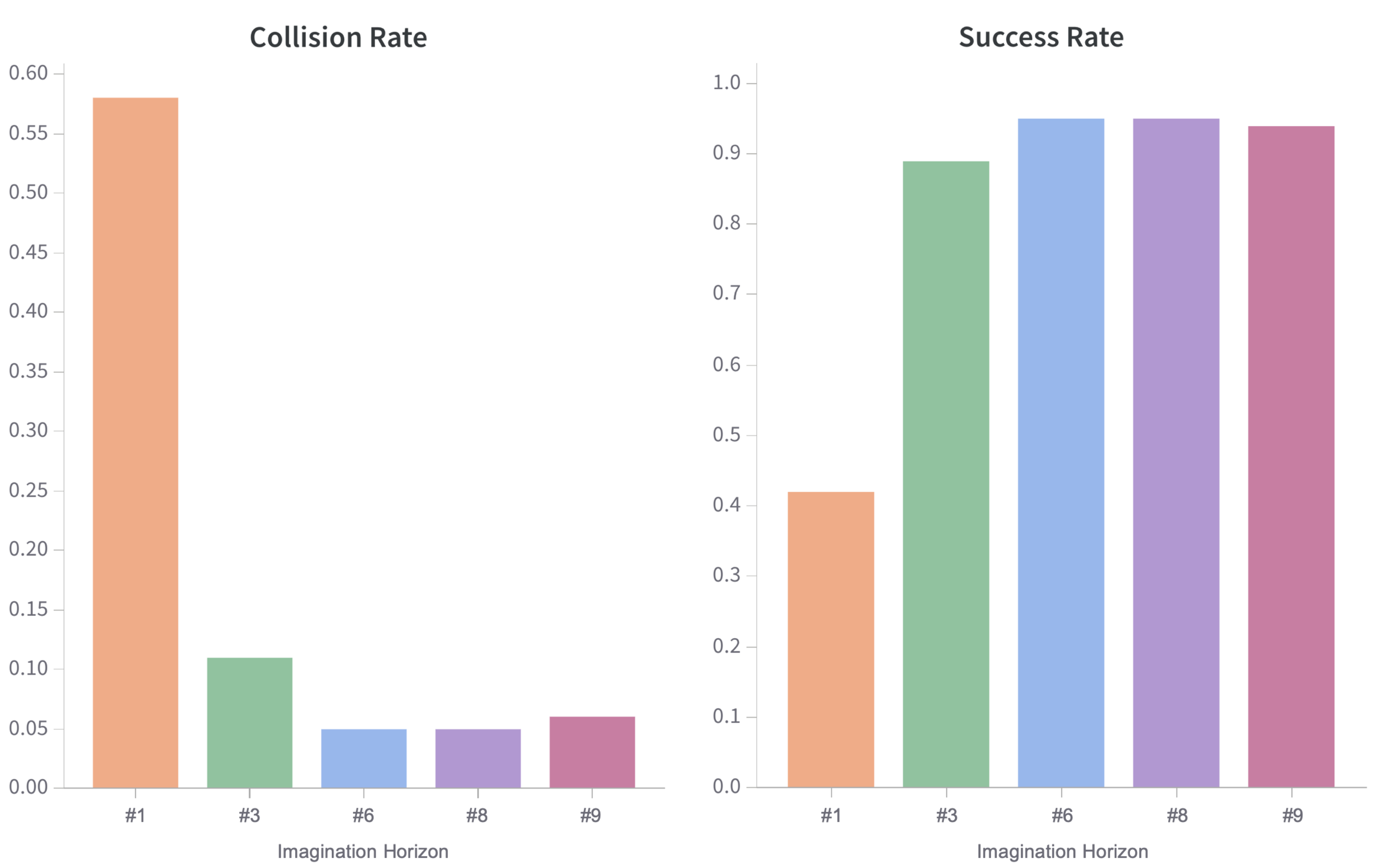}}
\caption{Performance comparison according to the imagination horizon. These results imply that imagined behaviors include critical information to infer interaction with surrounding objects.}
\label{fig:imagination_horizon}
\end{center}
\vskip -0.1in
\end{figure}

\textit{\textbf{4) How does the imagination horizon affect the performance of IAHRL?}}
Lastly, we examined the effects of the imagination horizon on the performance of IAHRL. The imagination horizon, one of the most important hyperparameters in IAHRL, determines how far into the future the low-level policies imagine the behaviors of our agent and surrounding objects. Figure \ref{fig:imagination_horizon} describes the performance according to the imagination horizon on the four-way intersection task. Note that imagined behaviors with one horizon consist of only the current position $(x_t, y_t)$ without further predictions by the behavior planner or perception module. Our agents with one or three imagination horizons perform too reckless behaviors, causing high collision and low success rates. In contrast, we observed that our agent with more than six imagination horizon maintains high performance across all ranges. This indicates that the imagined behaviors of our agent and surrounding vehicles contribute to the performance of our algorithm in urban driving tasks. We would like to emphasize that these results support our hypothesis that imagined behaviors can provide important information to infer interactions with surrounding objects, which is required to handle real-world navigation tasks.

\section{Conclusion}
We propose a novel and general navigation algorithm called IAHRL that efficiently integrates imagination into HRL. The key idea behind IAHRL is that the high-level policy infers interaction with surrounding objects by interpreting behaviors imagined with the low-level policies. We also propose a new attention mechanism that is permutation-invariant to the order of surrounding objects and prioritizes our agent over them. To evaluate IAHRL, we introduce five complex urban driving tasks, which are among the most challenging real-world navigation tasks. Experimental results demonstrate that our hierarchical agent performs safety-aware behaviors and properly interacts with surrounding vehicles, obtaining higher success rates and lower average episode steps than baselines. There are several interesting avenues for future work, which are not considered in this paper. First, we plan to extend our algorithm to be compatible with multiple sensor modalities. Leveraging their complementary advantages can help our agent perform more safe and interactive behaviors. Second, we are interested in applying our work to other environments, such as rough terrain or populated indoor environments. Third, we will investigate how to design a high-level policy that can handle a variable number of high-level actions or detection objects. This allows IAHRL to be applied to a more diverse range of navigation scenarios. Finally, we will explore the benefits of learning termination conditions for each behavior. This enables an agent to perform more flexible behaviors by maintaining selected behaviors within variable next steps. We hope our work contributes to further research on developing scalable navigation algorithms.

\bibliographystyle{./IEEEtran}
\bibliography{./IEEEabrv,./IEEEexample, ./references}

\begin{IEEEbiography}[{\includegraphics[width=1in,height=1.25in,clip,keepaspectratio]{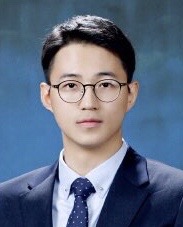}}]{Sang-Hyun Lee} received the B.S. degree in automotive engineering from Hanyang University, Seoul, South Korea, in 2015. He is currently pursuing the Ph.D. degree with the Department of Electrical Engineering and Computer Science, Seoul National University, Seoul, South Korea. He is also a Senior Researcher with ThorDrive, Seoul, South Korea. His current research areas include reinforcement learning, imitation learning, robot learning, and autonomous driving.
\end{IEEEbiography}

\begin{IEEEbiography}[{\includegraphics[width=1in,height=1.25in,clip,keepaspectratio]{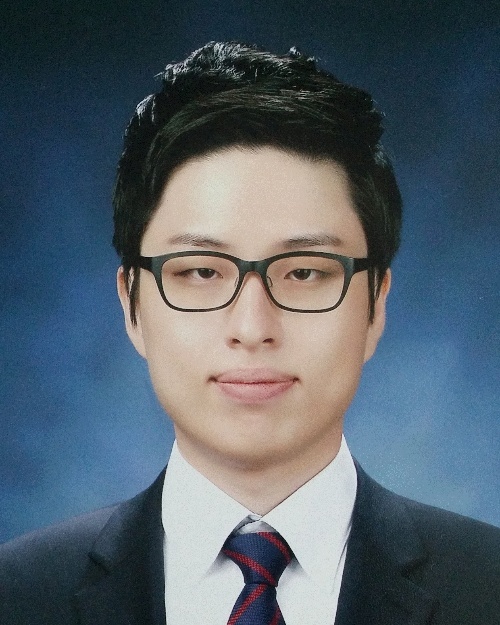}}]{Yoonjae Jung} received B.S. degree in electrical and computer engineering from Seoul National University, Seoul, South Korea in 2015 where he is currently pursuing Ph.D. degree. His current research areas include computer vision, deep learning and autonomous driving.
\end{IEEEbiography}

\begin{IEEEbiography}[{\includegraphics[width=1in,height=1.25in,clip,keepaspectratio]{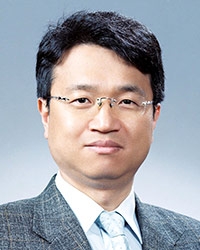}}]{Seung-Woo Seo} (Member, IEEE) received the B.S. and M.S. degrees in electrical engineering from Seoul National University, Seoul, South Korea, and the Ph.D. degree in electrical engineering from The Pennsylvania State University, University Park, PA, USA. In 1996, he joined as a Faculty Member with the School of Electrical Engineering, Institute of New Media and Communications and the Automation and Systems Research Institute, Seoul National University. He was a Faculty Member with the Department of Computer Science and Engineering, The Pennsylvania State University. He also served as a member of the Research Staff with the Department of Electrical Engineering, Princeton University, Princeton, NJ, USA. He is currently a Professor of electrical engineering with Seoul National University and the Director of the Intelligent Vehicle IT (IVIT) Research Center funded by the Korean Government and Automotive Industries.
\end{IEEEbiography}

\end{document}